\documentclass[11pt,a4paper]{article}
\usepackage{times}
\usepackage{latexsym}
\usepackage[T1]{fontenc}

\usepackage[acceptedWithA]{tacl2018v2}
\usepackage{url}
\usepackage{microtype}
\usepackage{graphicx}
\usepackage{tabularx}
\usepackage[font=small]{floatrow} 

\usepackage[utf8]{inputenc}
\usepackage{csquotes}

\usepackage{xcolor}

\usepackage{booktabs}
\usepackage{tabularx}

\usepackage{amsmath}
\usepackage{amssymb}
\usepackage{algorithm2e}
\usepackage{enumitem}

\newcommand{\YG}[2][]{\textcolor{brown}{#2}}

\newcommand{\bs}{\boldsymbol}

\hypersetup{
  pdftitle={Interactive Text Ranking with Bayesian Optimisation: 
A Case Study on Community QA and Summarisation},
  pdfauthor={CONFIDENTIAL}, 
}

\title{
Interactive Text Ranking 
with Bayesian Optimisation: 
A Case Study on Community QA and Summarisation
}

\author{Edwin Simpson$^{1,2}$ \hspace{2cm}
Yang Gao$^{1,3}$ \hspace{2cm}
Iryna Gurevych$^{1}$ \\
$^1$Ubiquitous Knowledge Processing Lab, Technische Universita\"t Darmstadt,\\
{\sf https://www.informatik.tu-darmstadt.de/} \\ 
$^2$Dept. of Computer Science, University of Bristol, {\sf edwin.simpson@bristol.ac.uk}\\
$^3$Dept. of Computer Science, Royal Holloway, University of London, {\sf yang.gao@rhul.ac.uk}\\
}

\begin{document}
\maketitle

\begin{abstract}
For many NLP applications, such as
question answering 
and summarisation, 
the goal is to select the best solution from a large space of candidates
to meet a particular user's needs.
To address the lack of user or task-specific training data,
we propose an interactive text ranking approach
that actively selects pairs of candidates, from which the user selects
 the best.
Unlike previous 
strategies, which attempt to learn a ranking across the whole candidate space, 
our method employs Bayesian optimisation to focus the user's 
labelling 
effort on high quality candidates 
and integrate prior knowledge 
to cope better with
small data scenarios. 
We apply our method to community question answering (cQA) and extractive multi-document summarisation,
finding that it significantly outperforms existing interactive  
approaches.
We also show that the ranking function learned by our 
method is an effective reward function for reinforcement learning,
which improves the state of the art for interactive summarisation.

\end{abstract}

\section{Introduction}
\label{sec:introduction}
Many text ranking tasks 
are highly 
specific to 
an individual user's topic of interest,
which presents a challenge for NLP systems that have 
not been trained to solve that user's problem.
Consider ranking summaries or answers to non-factoid questions:
a good solution requires understanding
the topic and the user's information needs~\citep{liu2008you,lopez1999using}.
%
We address this 
by proposing 
an interactive text ranking approach that efficiently
gathers user feedback and combines it with 
predictions from pretrained, generic models.

To minimise the amount of effort the user must expend to train a ranker,
we learn from pairwise preference labels, 
in which the user compares two candidates and labels the best one.
Pairwise preference labels can often be provided faster than ratings or class labels~\cite{yang2011ranking,kingsley2010preference,kendall1948rank},
can be used to rank candidates
using learning-to-rank~\cite{joachims2002optimizing}, preference learning~\cite{thurstone1927law}
or best-worst scaling~\cite{flynn2014best},
or to train a \emph{reinforcement learning (RL)} agent to find the optimal solution~\citep{wirth2017survey}.

To reduce the number of labels a user must provide, 
a common solution is \emph{active learning (AL)}.
AL learns a model by iteratively acquiring labels: 
at each iteration, it 
trains a model on the labels collected so far,
then uses an \emph{acquisition function}
to quantify the value of 
querying the user about a particular
pair of candidates. 
 The system then chooses the pairs with the highest values, and 
 instructs the user to label them.
The acquisition function implements one of many different strategies
to minimise the number of interaction rounds, 
such as reducing 
\textit{uncertainty}~\citep{settles2012active}
by choosing informative labels that help learn the model more quickly.

Many active learning strategies, 
such as the pairwise preference learning method of \citet{gao2018april},
aim to learn a good ranking model
 for all candidates, e.g., by querying the annotator about
candidates whose rank is most uncertain.
However, we often need  
to find and rank
only a small set of \emph{good} candidates to present to the user.
For instance, in question answering,
irrelevant answers should not be shown to the user,
so their precise ordering is unimportant 
and users should not waste time ranking them.
Therefore, 
by reducing uncertainty for all candidates, 
uncertainty-based 
AL strategies may waste
labels on sorting poor candidates.

Here, we propose 
an interactive method for ranking texts 
that replaces the standard uncertainty-based acquisition functions 
with acquisition functions for  
\emph{Bayesian optimisation (BO)}~\citep{movckus1975bayesian,brochu2010tutorial}.
In general, BO aims to find the maximum of a function
 while minimising the number of queries to an oracle.
Here, we use BO to maximise a ranking function that maps text documents to scores, treating the user 
as a noisy oracle.
Our BO active learning
strategy minimises the number of labels needed to find the best candidate,
in contrast to uncertainty-based strategies that
attempt to learn the entire ranking function.
This makes BO better suited to tasks such as question answering,
summarisation, or translation, where the aim is to find the best candidate and
those with low quality can simply be disregarded rather than ranked precisely. 
In this paper, we define two BO acquisition functions for interactive text ranking.

While our approach is designed to adapt a model to a highly specialised task,
generic models can provide hints to help us avoid low-quality candidates.
Therefore, we learn the ranking function itself
using a Bayesian approach, which integrates 
prior predictions from a 
generic model that is not tailored to the user.
Previous interactive text ranking
methods either do not exploit prior information ~\citep{baldridge2004active,avinesh2017joint,lin2017active,siddhant2018deep},  combine heuristics with user feedback 
 after active learning is complete~\citep{gao2018april},
or require expensive re-training of a non-Bayesian method~\cite{peris2018active}.
Here, we show how BO can use prior information to
 expedite interactive text ranking.
The interactive learning process is shown
 in Algorithm \ref{al:al_steps} and 
 examples of our system outputs are shown in Figures 
\ref{fig:example_cqa} and \ref{fig:example_sum}.
\begin{algorithm}
      \KwIn{ 
      candidate texts $\bs x$ with feature vectors
      $\phi(\bs x)$}
      \nl Initialise ranking model $m$\;
      \nl Set the training data $\bs D = \varnothing$\;
      \While{$| \bs D | < max\_interactions$}
      {
      \nl For each pair of texts $(x_a,x_b)$ in $\bs x$,
      compute $v = acquisition(\phi(x_a), \phi(x_b), \bs D, m)$\;
      \nl Set $\bs P_{i}$ to be the set of $batch\_size$ pairs 
      with the highest values of $v$\;    
      \nl Obtain labels $\bs y_{i}$ from user
      for pairs $\bs P_{i}$ \;
      \nl Add $\bs y_{i}$ and $\bs P_{i}$ to $\bs D$ \;
      \nl Train model $m$ on the training set $\bs D$ \;
      }
      \KwOut{ Return the trained model $m$ and/or 
      its final ranked list of candidate texts in $\bs x$. }
      \DontPrintSemicolon\;
     \caption{Interactive text ranking process with preference learning.}
     \label{al:al_steps}
    \label{fig:my_label}
\end{algorithm}
\begin{figure}
    \footnotesize 
    \noindent\fbox{
    \parbox{0.95\textwidth}{
    \textbf{Q:} \textit{Does whiskey go bad by freezing?}\\
    \textbf{A1:} It is to cool it down without dilluting it--ice cubes would melt. And yes, you could simply cool the entire bottle, but it wouldn't look that fancy. Note that some purists would wrinkle their noses and insist that whisky is best enjoyed at room temperature and perhaps with a small dash of spring water. And I'm soooo not going into a whisky vs. whiskey debate here.\\
    \textbf{A2:} \textbf{Putting strong spirits in the freezer should not harm them.} The solubility of air gases increases at low temperature, which is why you see bubbles as it warms up. \textbf{Drinks with a lower alcohol content will be affected in the freezer.} There is potential to freeze water out of anything with an alcohol content of 28\% or lower. Many people use the freezer to increase the alcohol content of their home brew in UK, by freezing water out of it--the alcohol stays in the liquid portion.
    \caption{Example from the Stack Exchange Cooking topic.
    Candidate answer A1 selected without user interaction 
    by COALA~\citep{ruckle2019coala}; 
    A2 chosen by our system (GPPL with IMP) after 10 user interaction. A2 answers the question (boldfaced texts)
    but A1 fails.
        \label{fig:example_cqa}
    }
    }
    }
\end{figure}
\begin{figure*}[t]
    \footnotesize
    \noindent\fbox{
    \parbox{0.97\columnwidth}{
        \textbf{(a):} 
        \textcolor{green}{A third leading advocate of the China Democracy Party who has been in custody for a month, Wang Youcai, was accused of ``inciting the overthrow of the government,''} the Hong Kong-based Information Center of Human Rights and Democratic Movement in China reported.
        China's central government \textcolor{blue}{ordered the arrest of a prominent democracy campaigner} and may use his contacts with exiled Chinese dissidents to charge him with harming national security, a colleague said Wednesday.
        One leader of a suppressed new political party will be tried on Dec. 17 on a charge of colluding with foreign enemies of China ``to incite the subversion of state power,'' according to court documents given to his wife on Monday.
           }
    }
    \noindent\fbox{
    \parbox{0.97\columnwidth}{ 
        \textbf{(b):} 
        \textcolor{orange}{The arrests of Xu and Qin} at their homes Monday night and the accusations against them and \textcolor{green}{Wang} were the sharpest action Chinese leaders have taken since dissidents began pushing to set up and legally register the \textcolor{orange}{China Democracy Party} in June.
        Hours before \textcolor{red}{China was expected to sign a key U.N. human rights treaty} and host British Prime Minister Tony Blair, \textcolor{blue}{police hauled a prominent human rights campaigner in for questioning Monday}.
        \textcolor{brown}{With attorneys locked up, harassed or plain scared, two prominent dissidents will defend themselves against charges of subversion} Thursday in China's highest-profile dissident trials in two years.
        Wang was a student leader in the 1989 Tiananmen Square democracy demonstrations.
         }
    }
    \noindent\fbox{
       \parbox{0.97\columnwidth}{
        \textbf{(c):}
        On the eve of \textcolor{red}{China's signing the International Covenant of Civil and Political Rights (ICCPR)} in October 1998, \textcolor{blue}{police detained Chinese human rights advocate Qin Yongmin for questioning}.
        Eight weeks after signing the ICCPR, Chinese police 
        \textcolor{orange}{arrested Qin and an associate in the China Democracy Party (CDP), Xu Wenli, without stating charges}.
        \textcolor{green}{Another CDP leader already in custody, Wang Youcai, was accused of "inciting the overthrow of the government".}
        Qin and Wang went to trial in December for inciting subversion.
        \textcolor{brown}{Police pressure on potential defense attorneys forced the accused to mount their own defenses}.
        Xu Wenli had not yet been charged.
    }
    }
    \caption{Example summaries for DUC'04 
     produced by RL (see Section \ref{sec:expts_rl})
    with a reward function learnt from 100 user interactions 
    using (a) the BT, UNC method of ~\citet{gao2018april}
    and (b) our GPPL, IMP method.
    (c) is a model summary written by an expert. 
   Each colour indicates a particular news event or topic, showing where it
   occurs in each summary.
   Compared to (a), summary (b) covers more of the events discussed in the reference, (c).
    \label{fig:example_sum}
    }
\end{figure*}

Our contributions are (1)
a Bayesian optimisation methodology for interactive text ranking 
that integrates prior predictions with
user feedback, 
(2) acquisition functions for Bayesian optimisation with 
pairwise labels,
and (3) 
empirical evaluations on community question answering (cQA) 
and extractive multi-document summarisation, 
which show that our method brings 
substantial improvements in ranking and summarisation performance
(e.g. for cQA, an average 25\% increase in answer selection accuracy over the next-best method with 10 rounds of user interaction).
We release the complete experimental software for future work\footnote{
\url{https://github.com/UKPLab/tacl2020-interactive-ranking}}.

\section{Related Work}\label{sec:related_work}

\paragraph{Interactive Learning in NLP. }
Previous work has applied active learning to tasks involving
ranking or optimising generated text, including 
summarisation~\citep{avinesh2017joint},
visual question answering~\cite{lin2017active}, and
translation~\citep{peris2018active}.
For summarisation, 
\citet{sokolov2016stochastic},
\citet{lawrence2018improving}
and \citet{singh2019end},
train reinforcement learners by querying the user directly for rewards,
which requires in the order of $10^5$ interactions.
\citet{gao2018april} dramatically
reduce the number of user interactions to the order of 
to $10^2$ by using active learning to learn 
a reward function for RL, 
an approach proposed by \citet{lopes2009active}.
These previous works 
use \emph{uncertainty sampling} strategies, 
which query the user about the candidates with the most
uncertain rankings 
to try to learn all candidates'  rankings
with a high degree of confidence.
We instead propose to find good candidates using an  optimisation strategy.
\citet{siddhant2018deep} carried out a large empirical study of 
uncertainty sampling for sentence classification,
semantic role labelling and named entity recognition,
finding that exploiting model uncertainty estimates provided 
by Bayesian neural networks improved performance.
Our approach also exploits Bayesian uncertainty estimates.

\paragraph{BO for Preference Learning.}



Bayesian approaches using Gaussian 
processes (GPs) have previously been used to reduce errors in NLP tasks
involving 
sparse or noisy labels~\cite{ cohn2013modelling, beck2014joint},
making them well-suited to learning from user feedback.
\emph{Gaussian process preference learning (GPPL)}~\citep{chu2005preference} enables
GP inference with pairwise preference labels. 
\citet{simpson2018finding} introduced scalable inference   
 for GPPL using
stochastic variational inference 
(SVI)~\cite{hoffman2013stochastic},
which outperformed SVM and LSTM methods at
ranking arguments by convincingness. 
They included a study on active learning with pairwise labels,
but tested GPPL only with uncertainty sampling, not BO.
Here, we adapt GPPL to summarisation and cQA, show how to integrate prior predictions, and 
propose a BO framework for GPPL that facilitates interactive text ranking.

\citet{eric2008active} proposed a BO approach for 
 pairwise comparisons but applied the approach only 
to a material design use case with a very simple feature space.
\citet{gonzalez2017preferential} proposed alternative BO strategies
for 
pairwise preferences,
but their approach requires 
expensive sampling
to estimate the utilities,
which is too slow for an interactive setting.
\citet{yang2018bayesian} also propose BO
with pairwise preferences,
but again, inference is expensive,
the method is only tested 
with fewer than ten features, 
and it uses an inferior \emph{probability of improvement} strategy
(see ~\citet{snoek2012practical}).
Our GPPL-based framework permits much faster inference 
even when the input vector has more than 200 features, and hence
allows rapid selection of new pairs when querying users.

~\citet{ruder2017learning}
use BO to select training data for transfer learning in NLP tasks such 
as sentiment analysis, POS tagging, and parsing. However, unlike our interactive text ranking
approach, 
their work does not involve pairwise comparisons
and is not interactive, as the optimiser learns by training and evaluating a model on the selected data.
In summary, previous work has not yet devised BO strategies for GPPL 
or suitable alternatives for interactive text ranking.

\section{Background on Preference Learning}
\label{sec:pairwise}
%
Popular preference learning models assume that
users choose a candidate from a pair with probability 
$p$, where $p$ is a function of the candidates' \emph{utilities}~\cite{thurstone1927law}.
Utility is defined as the value of a candidate to the user, i.e., it
quantifies how well that instance meets their needs.
When candidates have similar utilities, the user's choice is close to random,
while pairs with very different utilities are labelled consistently.
Such models include the Bradley--Terry model (BT)~\cite{bradley1952rank,luce1959possible,plackett1975analysis},
and the Thurstone--Mosteller model~\cite{thurstone1927law,mosteller1951remarks}. 

BT defines the probability that
candidate $a$ is preferred to candidate $b$ as follows:
\begin{flalign}
    & p(y_{a,b}) \!=\! \left(1 \!+\! \exp\left(\bs w^T\!\phi(a) \! - \! \bs w^T\!\phi(b)\right)\right)^{-1} &
\label{eq:bt}
\end{flalign}
where
$y_{a,b}=a \succ b$  is a binary preference label,
 $\phi(a)$ is the feature vector of $a$ and 
$\bs w^T$ is a weight parameter that must be learned.
To learn the weights, we treat
each pairwise label as two data points:
the first point has input features $\bs x = \phi(a)-\phi(b)$
and label $y$, 
and the second point is the reverse pair,
with $\bs x = \phi(b) - \phi(a)$ and label $1-y$. 
Then, we use standard techniques for logistic regression
to find the weights $\bs w$ that
minimise the L2-regularised cross entropy loss.
The resulting linear model can be used to predict labels for any unseen pairs, 
or to estimate candidate utilities, $f_a = w^T\phi(a)$, which can be used for
ranking.

\paragraph{Uncertainty (UNC).}
At each active learning iteration, the learner 
requests training labels for candidates that maximise 
the acquisition function. 
\citet{avinesh2017joint} proposed an 
\emph{uncertainty sampling} acquisition function
for interactive document summarisation, which 
defines the uncertainty about a single candidate's utility, $u$, as follows:
\begin{equation}
\label{eq:unc}
    u(a | \boldsymbol{D}) = \!\! \begin{cases} 
    p(a | \boldsymbol{D}) & \!\!\!\!\mbox{if  }\; p(a | \boldsymbol{D}) \leq 0.5 \\
    1\!-\!p(a | \boldsymbol{D}) & \!\!\!\!\mbox{if  }\; p(a | \boldsymbol{D}) > 0.5,
    \end{cases}
\end{equation}
where $p(a | \boldsymbol{D}) = ( 1 + \exp( -f_a ) )^{-1}$
is the probability that $a$ is a good candidate
and $\boldsymbol w$ is the set of BT model weights 
trained on the data collected so far, $\boldsymbol{D}$,
which consists of pairs of candidate texts and pairwise preference labels.
For pairwise labels, \citet{gao2018april}
define an acquisition function, which we refer to here as \emph{UNC}, 
which 
selects the pair of candidates $(a,b)$ with the two highest values of $u(a| \bs D)$ and $u(b| \bs D)$.

UNC is intended to focus labelling effort on candidates whose utilities are uncertain. 
If the learner is uncertain about whether candidate $a$ is a 
good candidate, $p(a | \boldsymbol{D})$ will be close to 0.5, so $a$ will 
have a higher chance of being selected. 
Unfortunately, it is also possible for $p(a | \boldsymbol{D})$ 
to be close to 0.5
even if $a$ has been labelled many times
if $a$ is a candidate of intermediate utility.
Therefore, when using UNC, 
labelling effort may be wasted re-labelling mediocre candidates.

The problem is that BT cannot distinguish two types of uncertainty.
The first is \emph{aleatoric} uncertainty due to the inherent unpredictability
 of the phenomenon we 
wish to model~\cite{senge2014reliable}. 
For example, when predicting the outcome of a coin toss, we model the outcome as random.
Similarly, given two equally preferable items,
we assume that the user assigns a preference label randomly. 
%
It does not matter how much training data we observe:
 if the items are equally good, 
 we are uncertain 
which one the user will choose.

The second type is
\emph{epistemic} uncertainty due to our lack of knowledge,
which can be reduced by acquiring more training
data, as this helps us to learn the  model's parameters with higher confidence.
BT does not quantify aleatoric and epistemic uncertainty separately, 
unlike Bayesian approaches~\citep{jaynes2003probability},
so 
we may repeatedly select items with similar utilities that do not require more labels.
To rectify this shortcoming, we replace
BT with a Bayesian model that both estimates the utility of a candidate
and quantifies the epistemic uncertainty in that estimate.

\paragraph{Gaussian Process Preference Learning}

Since BT does not quantify 
epistemic uncertainty in the utilities,
we turn to a Bayesian approach, GPPL.
GPPL uses a Gaussian process (GP)
to provide a nonlinear mapping from document feature vectors to utilities,
and assumes a Thurstone--Mosteller model for the pairwise preference labels.
Whereas BT simply estimates a scalar value of $f_a$ for 
each candidate, $a$,
GPPL outputs a posterior distribution over the utilities, $\boldsymbol f$,
of all candidate texts, $\boldsymbol{x}$:
\begin{equation}
    p( \boldsymbol{f} | \boldsymbol{\phi(x)}, \boldsymbol{D} ) = \mathcal{N}( \boldsymbol{\hat{f}}, \boldsymbol{C} ), 
\end{equation}
where 
$\boldsymbol{\hat{f}}$ is a vector of posterior mean utilities
and $\boldsymbol{C}$ is the posterior covariance matrix of the utilities.
The entries of $\boldsymbol{\hat{f}}$ are  
predictions of $f_a$ for each candidate given $\bs D$, 
and the diagonal entries of $\boldsymbol{C}$ represent posterior variance, which can be used to quantify 
uncertainty in the predictions. Thus, 
GPPL provides a way to separate candidates 
with uncertain utility
from those with middling utility but many pairwise labels.
In this paper, we
infer the posterior distribution over the utilities using
the scalable SVI method detailed by \citet{simpson20scalable}.




\section{Interactive Learning with GPPL}\label{sec:apl}

We now define four acquisition functions for GPPL 
that take advantage of the posterior covariance, 
$\boldsymbol{C}$, to account for uncertainty in the utilities.
Table~\ref{tab:acquisition_functions} summarises
these acquisition functions.
\begin{table*}
\footnotesize
 \setlength{\tabcolsep}{3pt}
 \centering
 \begin{tabularx}{\textwidth}{lXXXXXll}
   \toprule
   Learner & BT & BT & GPPL & GPPL & GPPL& GPPL & GPPL \\
   Strategy   & random & UNC & random & UNPA & EIG & IMP & TP \\
   \midrule
Considers epistemic uncertainty & N & Y & N & Y & Y & Y & Y \\
Ignores aleatoric uncertainty & N & N & N & N & Y & Y & Y \\
Supports warm-start & N & N & Y & Y & Y & Y & Y \\
Focus on finding best candidate & N & N & N & N & N & Y (greedy) & Y (balanced) \\
   \bottomrule
  \end{tabularx}
    \caption{Characteristics of active preference learning strategies. TP balances finding best candidate with exploration.}
    \label{tab:acquisition_functions}
\end{table*}




\paragraph{Pairwise Uncertainty (UNPA).} 

Here we propose a new adaptation of uncertainty sampling
to pairwise labelling with the GPPL model.
Rather than evaluating each candidate individually, as in UNC, 
we select the pair whose label is most uncertain.
UNPA selects the pair with label probability 
$p(y_{a,b})$ closest to $0.5$,
where, for GPPL:
\begin{flalign}
p(y_{a,b}) &= \Phi\left(\frac{\hat{f}_{a} - \hat{f}_{b}}{\sqrt{1 + v}} \right), & \\
v &= \boldsymbol{C}_{a,a}+\boldsymbol{C}_{b,b}-2\boldsymbol{C}_{a,b}, & 
\end{flalign}
where $\Phi$ is the probit likelihood and $\hat{f}_a$ is the posterior mean 
utility for candidate $a$.
Through $\boldsymbol{C}$, this function accounts for correlations between candidates' utilities and epistemic
uncertainty in the utilities.
However, 
for two items with similar expected utilities, $\hat{f}_a$ and $\hat{f}_b$,
the $p(y_{a,b})$ is close to $0.5$,
i.e., it has high aleatoric uncertainty.
Therefore, while UNPA will favour candidates with uncertain utilities,
it may still waste labelling effort on pairs with similar utilities 
but low uncertainty.


\paragraph{Expected Information Gain (EIG).}

We now define a second acquisition function for active learning with GPPL,
which has previously been adapted to GPPL by \citet{houlsby2011bayesian}
from an earlier information-theoretic strategy~\citep{mackay1992information}.
EIG
greedily reduces the epistemic uncertainty in the GPPL model by choosing
pairs that maximise 
\emph{information gain}, which
quantifies the information a pairwise label provides about 
the utilities, $\bs f$. 
Unlike UNPA, this function avoids pairs that only have 
high aleatoric uncertainty.
The information gain for a pairwise label,
$y_{a,b}$, is the reduction in entropy of the
distribution over the utilities, $\boldsymbol{f}$,
given $y_{a,b}$. 
\citet{houlsby2011bayesian} note that this can be more easily 
computed if it is reversed using a method known as
Bayesian 
active learning by disagreement (BALD),
which computes the reduction in entropy
of the label's distribution
given $\boldsymbol{f}$.
Since we do not know the value of $\boldsymbol{f}$,
we take the \emph{expected} information gain 
$\boldsymbol{I}$ with respect to $\boldsymbol{f}$:
\begin{flalign}
\mathrm{I}(y_{a,b},\boldsymbol{f}; \boldsymbol{D}) & = 
\mathrm{H}( y_{a, b} | \boldsymbol{D} ) - \mathbb{E}_{\boldsymbol{f}|\boldsymbol{D}}[ \mathrm{H}( y_{a, b} | \boldsymbol{f} ) ], 
&
\label{eq:EIG}
\end{flalign}
where $\mathrm{H}$ is Shannon entropy. 
Unlike the related \emph{pure exploration} strategy~\citep{gonzalez2017preferential}, 
Equation \ref{eq:EIG} can be computed in closed form, 
so does not need expensive sampling. 

\paragraph{Expected Improvement (IMP).}
The previous acquisition functions for AL are uncertainty-based,
and spread labelling effort across all items whose utilities are uncertain.
However, for tasks such as summarisation or 
cQA, the goal is to
find the best candidates.
While it is important to distinguish between good and optimal candidates,
 it is wasted effort to compare candidates that we are already 
confident are not the best, even if their utilities are still uncertain.
We propose to address this using an
acquisition function for Bayesian optimisation (BO)
that estimates the \emph{expected improvement}~\citep{movckus1975bayesian}
of a candidate, $a$, over our current estimated best solution, $b$,
given current pairwise labels, $\boldsymbol{D}$.
Here, we provide the first closed-form acquisition function that uses 
expected improvement for pairwise preference learning.

We define \emph{improvement} as the quantity
$\max\{0, f_{a} - f_{b}\}$, where $b$ is our current best item and $a$ is our new
candidate.
Since the values of $f_a$ and $f_b$ are uncertain,
we compute the \emph{expected} 
improvement as follows.
First, we estimate the posterior distribution over the candidates' utilities, 
$\mathcal{N}(\hat{\bs f}, \bs C)$,
then find the current best utility:
$\hat{f}_{b} = \max_{i}\{\hat{f}_{i}\}$.
For any candidate $a$, the difference $f_a - f_{b}$ 
is Gaussian-distributed
as it is a sum of Gaussians.  
The probability that this is larger than 
zero is given by the cumulative density function, 
$\Phi(z)$, where $z = \frac{\hat{f}_{a} - \hat{f}_{b}}{\sqrt{v}}$.
We use this
to derive expected improvement, which results in 
the following closed form equation:
\begin{flalign}
\mathrm{Imp}(a; \boldsymbol{D}) &= \sqrt{v}z\Phi(z) + \sqrt{v}\mathcal{N}(z; 0,1) , &
\end{flalign}
This weights the probability of finding a better solution, $\Phi(z)$, 
by the amount of improvement, $\sqrt{v}z$.
Both terms account for 
how close $\hat{f}_a$ is to $\hat{f}_{b}$ 
through $z$,
as a larger distance causes $z$ to be more negative, 
which decreases both the probability $\Phi(z)$ and the density  $\mathcal{N}(z; 0,1)$.
Expected improvement also accounts for the uncertainty 
in both utilities through the posterior standard deviation, 
$\sqrt{v}$, which scales both terms.
 All candidates have positive expected improvement, as there is a non-zero probability
 that labelling them will lead to a new best item;
 otherwise, the current best candidate remains,
 and improvement is zero.

To select pairs of items, the IMP strategy greedily chooses the current best
item and the item with the greatest expected improvement. 
Through the consideration of posterior uncertainty, IMP
trades-off \emph{exploration} of unknown candidates
with \emph{exploitation} of promising candidates. In contrast, 
uncertainty-based strategies are pure exploration. 

\paragraph{Thompson Sampling with Pairwise Labels (TP).}

Expected improvement is known to over-exploit
in some cases~\citep{qin2017improving}:
It chooses where to sample based on the current 
distribution, so if this distribution underestimates the mean and variance 
of the optimum, it may never be sampled.
To increase exploration, we propose a strategy that uses
Thompson sampling~\citep{thompson1933likelihood}. 
Unlike IMP, which is deterministic,
TP introduces random exploration through sampling.
TP is similar to dueling-Thompson sampling for  continuous input domains~\citep{gonzalez2017preferential}, 
but uses an information gain step (described below) and
samples from a pool of discrete candidates.

We select an item using Thompson sampling as follows: first 
draw a sample of candidate utilities from their posterior distribution, $\boldsymbol{f}_{\mathrm{thom}} \sim \mathcal{N}(\boldsymbol{\hat{f}}, \boldsymbol{C})$,
then choose the item $b$ with the highest score in the sample.
This sampling step depends on a Bayesian approach to provide a posterior distribution from which to sample.
Sampling means that while candidates with
high expected utilities have higher values of 
$f_{\mathrm{thom}}$ in most samples,
 other candidates may also have the highest score in some samples. 
 As the number of samples $\to \infty$, the
number of times each candidate is chosen is 
proportional to the probability that it is the best candidate.

To create a pair of items for preference learning,
the TP strategy computes the expected information gain for all pairs that include 
candidate $b$,
and chooses the pair with the maximum. 
This strategy is less greedy than IMP as it 
allows more learning about uncertain items through both the Thompson sampling step and the information gain step. However, compared to EIG, the first step focuses effort more on items with potentially high
scores.

\paragraph{Using Priors to Address Cold Start.}

In previous work on summarisation~\citep{gao2018april}, 
the BT model was trained from a \emph{cold start}, i.e., with no
prior knowledge or pretraining.
Then, after active learning was complete, 
the predictions from the trained model were combined with 
prior predictions based on heuristics
by taking an average of the normalised scores from both methods.
We propose to use such
prior predictions to determine an \emph{informative prior} for GPPL,
enabling the active learner 
to make more informed choices of candidates to label
at the start of the active learning process,
thereby alleviating the cold-start problem.

We integrate pre-computed predictions
 as follows.
Given a set of prior predictions, $\boldsymbol{\mu}$, from a heuristic or pre-trained model,
we set the prior mean of the Gaussian process to $\boldsymbol{\mu}$
before collecting any data,
so that the candidate utilities have the prior
$p( \boldsymbol{f} | \boldsymbol{\phi(x)} ) = 
    \mathcal{N}( \boldsymbol{\mu}, \boldsymbol{K} )$,
where $\boldsymbol{K}$ is a hyper-parameter.
Given this setup, AL can now 
take the prior predictions into account when choosing pairs of candidates for labelling. 




\section{Experiments}\label{sec:expts}

We perform experiments on three tasks to 
test our interactive text ranking approach:
(1) 
Community question answering (cQA)
-- identify the best answer to a given question from a pool of candidate answers;
(2) Rating extractive multi-document summaries
according to a user's preferences;
(3) Generating an 
extractive multi-document summary by training a reinforcement learner with
the ranking function from 2 as a reward function.
Using interactive learning to learn the reward function rather than the policy
reduces the number of user interactions from 
many thousands to 100 or less.
These tasks involve highly specialised questions or topics
where generic models could be improved with user feedback.
For the first two tasks, we simulate the interactive process in Algorithm \ref{al:al_steps}.
The final task uses the results of this process.

\paragraph{Datasets.}
Both the cQA and multi-document summarisation 
datasets were chosen because the 
answers and candidate summaries in these datasets
are multi-sentence documents that
take longer for users to read compared to tasks such 
as factoid question-answering. 
We expect our methods to have the greatest impact 
in this type of long-answer scenario by
minimising user interaction time.

For cQA, 
we use datasets consisting of questions posted on StackExchange
in the communities \emph{Apple}, \emph{Cooking} and \emph{Travel},
along with their accepted answers and candidate answers taken from related questions~\citep{ruckle2019coala}.
Each accepted answer was marked by the user who posted the question,
so reflects that user's own opinion.
Dataset statistics are shown in Table \ref{tab:datasets}. 
\begin{table}
 \centering 
 \small
 \begin{tabularx}{\columnwidth}{lXXX}
   \toprule
      cQA Topics & \#questions & \#accepted answers & \#candidate answers \\
   \midrule
   Apple    & 1,250 & 1,250 & 125,000 \\
   Cooking  & 792 & 792 & 79,200 \\
   Travel   & 766 & 766 & 76,600 \\
   \toprule
      Summarisation & \#topics & \#model & \#docs \\
      Datasets & & summaries \\
   \midrule
   DUC 2001 & 30 & 90 & 300 \\
   DUC 2002 & 59 & 177 & 567 \\
   DUC 2004 & 50 & 150 & 500 \\
   \bottomrule
  \end{tabularx}
    \caption{Dataset statistics for summarisation and cQA.}
    \label{tab:datasets}
\end{table}

For summarisation, we use the DUC datasets\footnote{\url{http://duc.nist.gov/}}.
which contain model summaries written by experts for collections of documents related to a 
narrow topic.
Each topic has three model summaries, each written by a different expert
and therefore reflecting different opinions
about what constitutes a good summary.
Compared to single-document summarisation,
the challenging multi-document case
is an ideal testbed for interactive approaches, since the diversity 
of themes within a collection of documents 
makes it difficult to identify a single, concise summary suitable for all users.

\paragraph{Priors and Input Vectors.}

We use our interactive approach to improve a set of prior
predictions provided by a pretrained method.
For cQA, we first choose the previous state-of-the-art for long answer selection,
\emph{COALA}~\citep{ruckle2019coala},
which estimates the relevance of answers to a question by
extracting \emph{aspects} (e.g., n-grams or syntactic structures) 
from the question and answer texts using CNNs,
then matching and aggregating the aspects.
For each topic, we train an instance of COALA on the training split
given by \citet{ruckle2019coala}, 
then run the interactive process on the test set, 
i.e., the dataset in Table \ref{tab:datasets},
to simulate a user interactively refining the answers selected for their question. 
As inputs for the BT and GPPL models, we use the COALA feature vectors: 
for each question, COALA extracts aspects from the question and its candidate answers;
each dimension of an answer's 
50-dimensional 
feature vector 
encodes how well the answer covers one of the extracted aspects.

Next we apply our interactive approach to refine
predictions from the current state of the art~\citep{xu2019passage},
which we refer to as \emph{BERT-cQA}.
This method places two dense layers with 100 and 10 hidden units
 on top of BERT~\citep{devlin2019bert}. 
 As inputs to BERT, we concatenate the question and candidate answer and pad sequences to 512 tokens (4\% QA pairs are over-length and are truncated).
The whole model
is fine-tuned on the StackExchange training sets, the same 
as COALA. 
In our simulations, we use 
the fine-tuned, final-layer [CLS] embeddings with 768 dimensions 
as inputs to BT and GPPL for each question-answer pair. 

As prior predictions for summary ratings
we first evaluate \emph{REAPER}, 
a heuristic evaluation function described by
\citet{ryang2012framework}. 
We obtain \emph{bigram+} feature vectors for candidate summaries
by augmenting bag-of-bigram embeddings with additional features
proposed by \citet{rioux2014fear}. 
The first 200 dimensions of the feature vector
have binary values 
to indicate the presence of each of
the 200 most common bigrams in each topic
after tokenising, stemming and applying a stop-list.
The last 5 dimensions contain the following:
the fraction of the 200 most common bigrams 
that are present in the document (coverage ratio);
fraction of the 200 most common bigrams that occur more than
once in the document (redundancy ratio);
document length divided by 100 (length ratio);
the sum over all extracted sentences of the reciprocal of the position of the
extracted sentence in its source document (extracted sentence position feature);
a single bit to indicate if document length 
exceeds the length limit of 100 tokens.
The same features are used for both tasks (2) learning summary ratings and (3) reinforcement learning.

We also test 
prior predictions from a state-of-the-art summary scoring method, 
\emph{SUPERT}~\citep{gao2020supert},
which uses 
a variant of BERT that has been fine-tuned on news articles
 to obtain 1024-dimensional contextualised embeddings of a summary. 
To score a summary, 
SUPERT extracts a pseudo-reference summary from the source documents,
then compares its embedding with that of the test summary.
With the SUPERT priors we compare bigram+ feature vectors and the SUPERT embeddings as input to BT and GPPL for task (2).

\paragraph{Interactive Methods. }
As baselines, we 
test BT as our preference learner
with random selection 
and the UNC active learning strategy,
and GPPL as the learner with 
random selection.
We also combine GPPL with the
acquisition functions described in Section \ref{sec:apl},
UNPA, EIG, IMP and TP. 
For random sampling, we repeat each experiment ten times.

\paragraph{Simulated Users. }
In tasks (1) and (2), we simulate a user's preferences
with a noisy oracle based on the user-response models of ~\citet{viappiani2010optimal}.
Given gold standard scores for two documents, $g_a$ and $g_b$,
the noisy oracle prefers document $a$ with probability 
$p(y_{a,b} | g_a, g_b) = (1+\exp( \frac{g_b-g_a}{t}))^{-1}$,
where $t$ is a parameter that controls the noise level.
Both datasets contain
model summaries or gold answers, but no
gold standard scores. 
We therefore estimate gold scores by computing a ROUGE score
of the candidate summary or answer, $a$, against the model summary or gold answer, $m$. 
For cQA, we take the ROUGE-L score as a gold score,  
as it is a well-established metric
for evaluating question answering systems (e.g. 
~\citet{nguyen2016ms,bauer2018commonsense,indurthi2018cut})
and set $t=0.3$, which results in annotation accuracy of 83\% 
(the fraction of times
  the pairwise label corresponds to the gold ranking).

For summarisation, 
we use $t=1$, 
which gives noisier annotations with 66\% accuracy,
reflecting the greater difficulty of choosing between two summaries.
This corresponds to accuracies of annotators
found by \citet{gao2019preference} 
when comparing summaries from the same datasets.
As gold for summarisation,
we combine ROUGE scores using the following formula,
previously found to correlate well with human preferences
on a comparable summarisation task~\citep{avinesh2017joint}:
\begin{flalign}
g_a & \approx R_{comb} = \frac{\mathrm{ROUGE_1}(a, m)}{0.47} +  \nonumber \\
& \frac{\mathrm{ROUGE_2}(a,m)}{0.22} + \frac{\mathrm{ROUGE_{SU4}}(a,m)}{0.18}.
\label{eq:r_comb}
\end{flalign}
Following \citet{gao2019preference}, 
we normalise the gold scores $g_a$ to the range $[0, 10]$.

\subsection{Warm-start Using Prior Information}

We compare two approaches to integrate the prior 
predictions of utilities
computed before acquiring user feedback.
As a baseline, \emph{sum} applies a weighted mean 
to combine the prior predictions with posterior predictions learned using GPPL or BT. 
Based on preliminary experiments,  
we weight the prior and posterior predictions equally.
\emph{Prior}
sets the prior mean of GPPL to the value of the prior predictions,
as described in Section \ref{sec:apl}.
Our hypothesis is that \emph{prior} will provide more information
at the start of the interactive learning process and help the learner to
select more informative pairs.


\begin{table}[t]
 \centering
 \small
  \setlength{\tabcolsep}{3pt}
  \begin{tabularx}{\columnwidth}{llXXX}
    \toprule
    Strategy & Prior &  \multicolumn{3}{c}{Datasets}  \\
    \midrule 
    \multicolumn{5}{c}{\textit{Accuracy for cQA with COALA priors}} \\
    \multicolumn{2}{l}{\textit{\#interactions=10}} & Apple & Cooking & Travel \\
    random & sum & .245 & .341 & .393 \\
    random & prior & \textbf{.352} & \textbf{.489} & \textbf{.556} \\
    UNPA & sum & \textbf{.293} & \textbf{.451} & .423 \\
    UNPA & prior & .290 & .392 & \textbf{.476} \\
    IMP & sum &  .373 & .469 & .466  \\
    IMP & prior & \textbf{.615} & \textbf{.750} & \textbf{.784} \\
    \midrule
    \multicolumn{5}{c}{\textit{NDCG@1\% for summarisation with REAPER priors}}\\
    \multicolumn{2}{l}{\emph{\#interactions=20}} & DUC'01 & DUC'02 & DUC'04 \\
    random & sum & \textbf{.595} & \textbf{.623} & \textbf{.652}\\
    random & prior & .562 & .590 & .600 \\
    UNPA & sum & .590 & .628 & \textbf{.650} \\
    UNPA & prior & \textbf{.592} & \textbf{.635} & .648 \\
    IMP & sum &  .618 &  .648 & .683 \\
    IMP & prior & \textbf{.654}& \textbf{.694} & \textbf{.702}\\
    \bottomrule
  \end{tabularx}
  \caption{
  The effect of integrating pre-computed predictions as Bayesian priors vs. taking a weighted mean of pre-computed and posterior predictions. 
  }
  \label{tab:warmstart}
\end{table}
Table \ref{tab:warmstart} presents results of a comparison on a subset of strategies,
showing that \emph{prior} results in higher performance
in many cases.
Based on the results of these experiments, we 
apply \emph{prior} to all further uses of GPPL.

\subsection{Community Question Answering} 


We hypothesise that the prior ranking given by COALA 
can be improved by incorporating a small amount of user feedback for each
question. 
Our interactive process 
aims to find the best answer to a specific question, 
rather than learning a model that transfers to new questions,
hence preferences are sampled for questions in 
 the \emph{test} splits. 

To evaluate the top-ranked answers from each method, 
we compute accuracy as the fraction of top answers that
match the gold answers.
We also compare the five highest-ranked solutions
to the gold answers 
using \emph{normalised discounted cumulative gain} (NDCG@5)
with ROUGE-L as the relevance score.
NDCG@k evaluates the relevance of the top $k$ ranked items,
putting more weight on higher-ranked items~\citep{jarvelin2002cumulated}.

The results in the top half of Table \ref{tab:cqa} show that with only 10 user interactions,
most methods are unable to improve performance over pre-trained COALA.
UNC, UNPA, EIG and TP are out-performed by random selection
and IMP
($p\ll .01$ using a two-tailed Wilcoxon signed-rank test).

\begin{figure}[t]
    \centering
    \includegraphics[width=\textwidth,clip=True,trim=10 5 40 40]{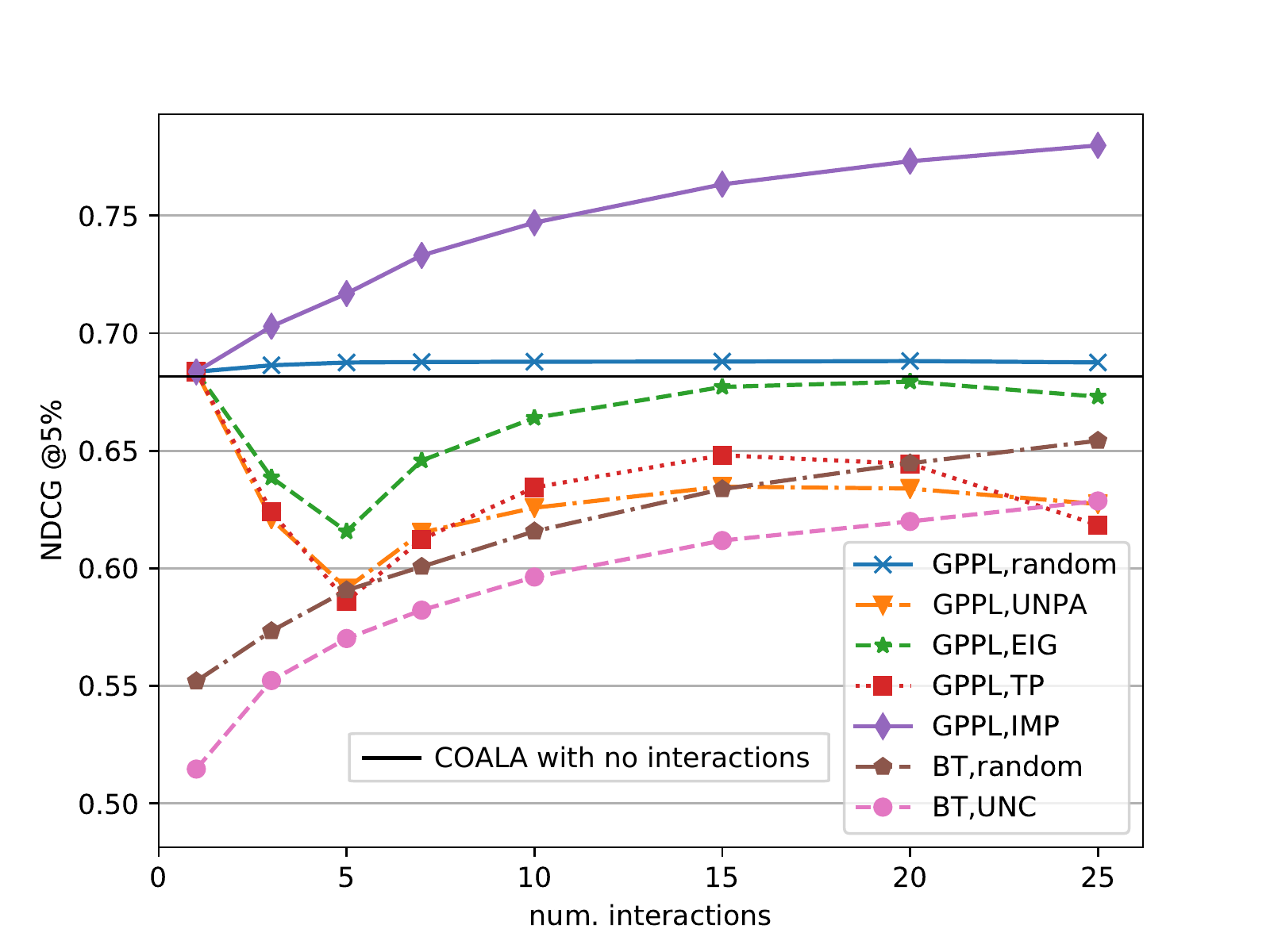}
    \caption{NDCG@5 with increasing
    interactions, COALA prior,
    mean across 3 cQA topics.
    }
    \label{fig:progress_coala}
\end{figure}
To see whether the methods improve given more feedback, 
Figure \ref{fig:progress_coala}
plots NDCG@5 against number of interactions.
While IMP performance increases substantially, 
random selection improves only very slowly. 
Early interactions cause a performance drop with UNPA, EIG, and
TP. 
This is unlikely to be caused by noise in the 
cQA data, since preference labels
are generated using ROUGE-L scores computed
against the gold answer.
The drop is because uncertainty-based methods 
initially sample many low-quality candidates with high uncertainty. 
This increases the predicted utility of the
preferred candidate in each pair, sometimes 
exceeding better candidates that were ranked higher by the prior,
pushing them out of the top five.
Performance rises once the uncertainty 
of mediocre candidates has
been reduced and stronger candidates are selected.
Both BT methods start from a
worse initial position but improve consistently, 
as their initial samples are not biased by the prior predictions,
although UNC remains worse than random.
\begin{table}[t]
 \centering
 \small
 \setlength{\tabcolsep}{3pt}
  \begin{tabularx}{\columnwidth}{llXXXXXX}
    \toprule
    & Strat & \multicolumn{2}{c}{Apple} & \multicolumn{2}{c}{Cooking} & \multicolumn{2}{c}{Travel} \\
    Learner & -egy & acc & N5 & acc & N5 & acc & N5 \\
    \midrule 
    \multicolumn{2}{l}{COALA} & .318 & .631 & .478 & .696 & .533 & .717 \\
    \midrule
    \multicolumn{8}{l}{\emph{COALA prior, \#interactions=10}}\\ 
    BT &  random & .272 & .589 & .368 & .614 & .410 & .644 \\
    BT &  UNC & .233 & .573 & .308 & .597 & .347 & .619 \\
    GPPL & random &  .352 & .642 & .489 & .699 & .556 & .722 \\
    GPPL & UNPA & .290 & .591 & .392 & .631 & .476 & .656 \\
    GPPL & EIG & .302 & .628 & .372 & .671	& .469 & .692 \\
    GPPL & TP & .274 & .592 & .353 & .636 & .414 & .675 \\
    GPPL & IMP & \textbf{.615} & \textbf{.714} & \textbf{.750} & \textbf{.753} & \textbf{.784} & \textbf{.774} \\
    \toprule
    \multicolumn{2}{l}{BERT-cQA} & .401 & .580 & .503 & .625 & .620 & .689  \\
    \midrule
    \multicolumn{8}{l}{\textit{BERT-cQA prior, \#interactions=10}}\\
    BT &  random & .170 & .626 & .228 & .637 & .315 & .676  \\
    BT &  UNC & .129 & .580 & .181 & .583 & .326 & .618 \\
    GPPL & random & .407 & .593 & .510 & .594 & .631 & .657   \\
    GPPL & EIG & .080 & .559 & .140 & .552 & .095 & .526 \\
    GPPL & IMP & \textbf{.614} & \textbf{.715} & \textbf{.722} & \textbf{.731} & \textbf{.792} & \textbf{.744} \\
    \bottomrule
  \end{tabularx}
  \caption{Interactive text ranking for cQA. N5=NDCG@5, acc=accuracy. 
  }
  \label{tab:cqa}
\end{table}
\begin{figure}[htbp]
    \centering
    \includegraphics[width=\textwidth,clip=True,trim=10 5 40 40]{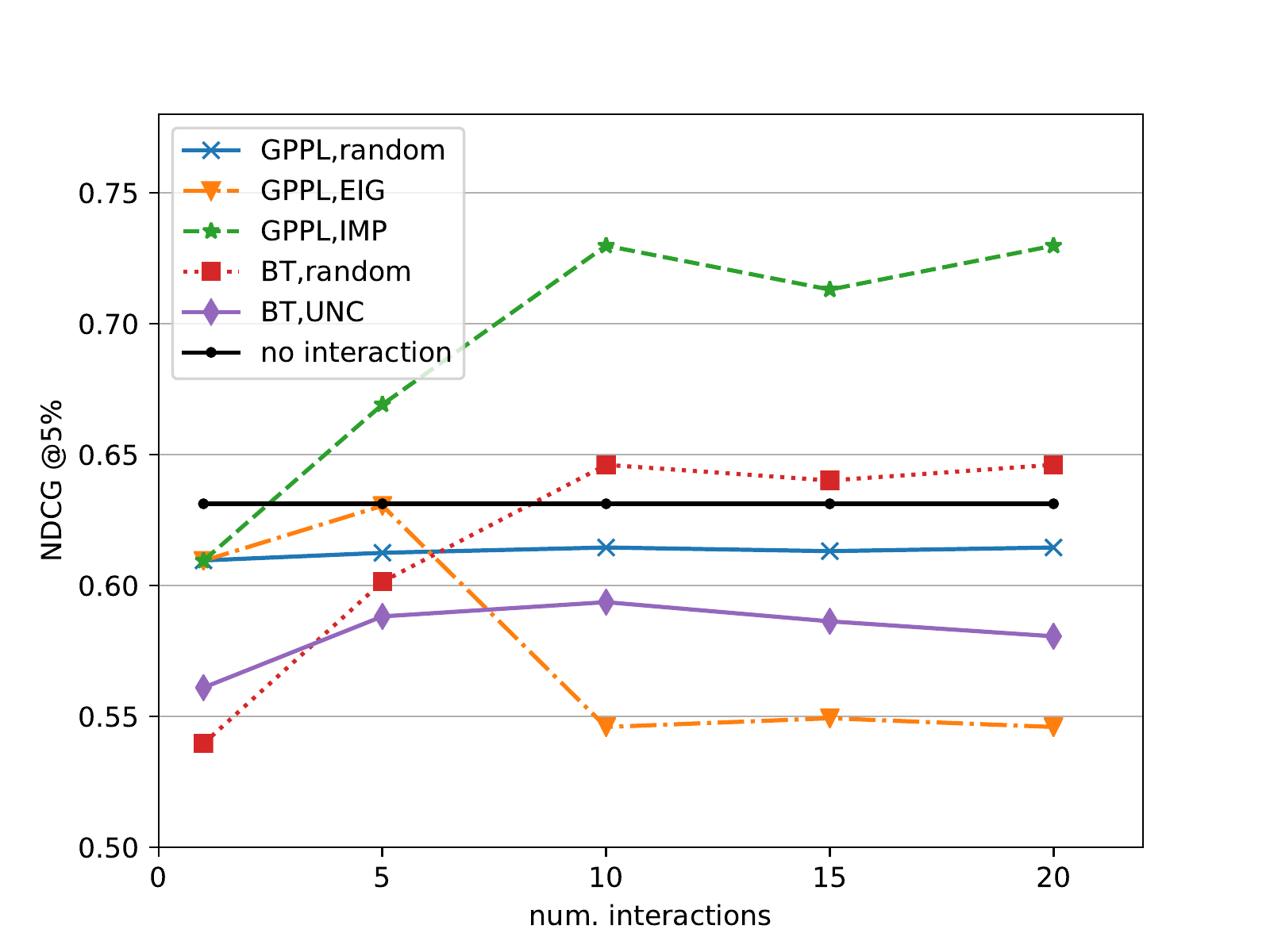}
    \caption{NDCG@5  with increasing number of interactions. BERT-cQA prior. 
    Mean across 3 cQA topics.
    }
    \label{fig:progress_bertcqa}
\end{figure}

The bottom half of Table \ref{tab:cqa} and Figure \ref{fig:progress_bertcqa}
show results for key methods with BERT-cQA priors and embeddings.
The initial predictions by BERT-cQA have higher accuracy than COALA
but lower NDCG@5.  
BERT-based models better account for 
question and answer semantics, 
leading to higher accuracy, 
but place less emphasis on lexical similarity, 
which reduces the ROUGE-L scores of 
top-ranked answers and consequently, NDCG@5.
While IMP remains the most successful method, the end result is not a clear improvement over
COALA, with a collapse in accuracy for the uncertainty-based EIG and both BT methods.
As with COALA, these uncertainty-based methods focus 
initially on middling candidates,
but due to the 
sparsity of the data with high-dimensional BERT-cQA embeddings,
more samples are required to reduce their uncertainty  
before these methods start to sample strong candidates. 
The flexibility of the GP model means that it is particularly affected by data
sparsity, hence the poor performance of EIG.

\subsection{Interactive Summary Rating}
\label{subsec:inter_summary_rating}
We apply interactive learning to refine a ranking 
over candidate summaries given prior information. 
For each topic, we create 10,000 candidate summaries
with fewer than 100 words each, which are constructed by 
uniformly selecting sentences at random from the input
documents.
To determine whether some strategies benefit from 
more samples, we test each active learning method
with between 10 and 100 user interactions
with noisy simulated users.
The method is fast enough for interactive scenarios:
on a standard Intel desktop workstation with
a quad-core CPU and no GPU, 
updates to GPPL at each iteration require around one second. 

We evaluate the quality of the
100 highest-ranked summaries using NDCG@1\%,
and compute the Pearson correlation, $r$, 
between the predicted utilities for all candidates
and the combined ROUGE scores (Eq. \YG[\ref]{\eqref{eq:r_comb}}). 
Unlike NDCG@1\%, $r$ 
does not focus on higher-ranked candidates but
considers the utilities for all candidates. 
Hence we do not expect that IMP or TP, which optimise
the highest-ranked candidates, 
will have the highest $r$.

\begin{figure}[t]
    \centering
    \includegraphics[width=\textwidth,clip=True,trim=8 5 35 40]{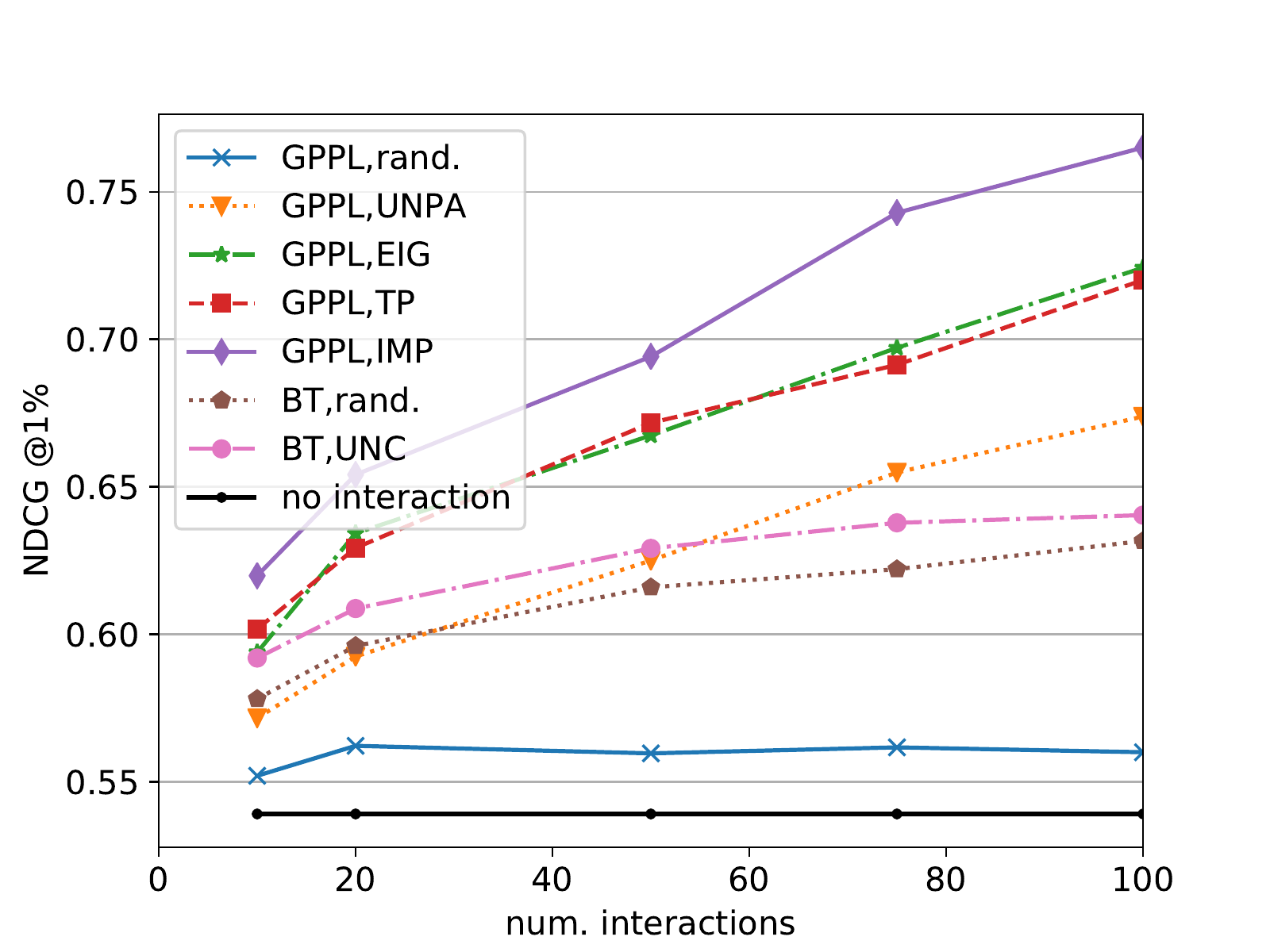}
    \caption{
    DUC'01, REAPER prior, bigram+ features, 
    changes in NDCG@1\% with increasing interactions.
    }
    \label{fig:progress_duc01}
\end{figure}
\begin{table}[h]
 \centering
 \small
  \setlength{\tabcolsep}{3pt}
  \begin{tabularx}{\columnwidth}{llXXXXXX}
    \toprule
    Lear & Strat & \multicolumn{2}{c}{DUC'01} & \multicolumn{2}{c}{DUC'02} & \multicolumn{2}{c}{DUC'04} \\
    -ner & -egy & N1 & r & N1 & r & N1 & r \\
    \midrule 
    \multicolumn{2}{l}{REAPER} & .539 & .262 & .573 & .278 & .597 & .322 \\
    \midrule
    \multicolumn{8}{l}{\emph{REAPER prior, bigram+ features, \#interactions=20}} \\
    BT & rand. &.596 & .335 & .626 & .358 & .659 & .408\\
    BT & UNC  & .609 & .340 & .641 & .365 & .674 & .415 \\
    GPPL & rand. &.558 & .248 & .590 & .266 & .603 & .289 \\
    GPPL & UNPA& .592 & .307 & .635 & .370 & .648 & .397 \\
    GPPL & EIG & .634 & .327 & .665 & .383 & .675 & .404 \\
    GPPL & TP & .629 & \textbf{.378} & .665 & \textbf{.403} & .690 & \textbf{.453}\\
    GPPL & IMP & \textbf{.654} & .303 & \textbf{.694} & .345 & \textbf{.702} & .364 \\
    \toprule 
    \multicolumn{2}{l}{SUPERT} & .602 & .382 & .624 & .400 & .657 & .438  \\
    \midrule
    \multicolumn{8}{l}{\emph{SUPERT prior, bigram+ features, \#interactions=20}} \\
    BT & rand. & .633 & .415 & .654 & \textbf{.438} & .684 & \textbf{.483} \\
    BT & UNC & .550 & .277 & .561 & .287 & .588 & .334 \\
    GPPL & rand. & .601 & .351 & .630 & .377 & .657 & .419 \\
    GPPL & EIG & .633 & .365 & .662 & .399 & .671 & .435\\
    GPPL & TP & .649 & \textbf{.417} & .668 & .437 & .698 & .479 \\
    GPPL & IMP & \textbf{.653} & .322 & \textbf{.696} & .374 & \textbf{.717} & .407\\
    \midrule
    \multicolumn{8}{l}{\emph{SUPERT prior, SUPERT embeddings, \#interact.=20}} \\
    GPPL & IMP & .624 & .297 & .630 & .284 & .653 & .339 \\   
    \midrule
    \multicolumn{8}{l}{\emph{SUPERT prior, bigram+ features, \#interactions=100}} \\ 
    GPPL & IMP & .668 & .308 & \textbf{.788} & .466 & \textbf{.815} & .521\\   
    \midrule
    \multicolumn{8}{l}{\emph{SUPERT prior, SUPERT embeddings, \#interact.=100}} \\ 
    BT & rand. & .661 & \textbf{.466} & .696 & \textbf{.504} & .727 & \textbf{.543} \\
    BT & UNC & .634 & .420 & .656 & .453 & .678 & .495  \\
    GPPL & rand. & .594 & .354 & .617 & .387 & .643 & .415 \\
    GPPL & EIG & .611 & .372 & .647 & .415 & .682 & .471 \\
    GPPL & IMP & \textbf{.728} & .376 & \textbf{.752} & .407 & \textbf{.769} & .447 \\    
    \bottomrule
  \end{tabularx}
  \caption{Interactive Summary Rating. N1=NDCG@1\%,  r=Pearson's correlation coefficient. Bold indicates best result for each prior and number of interactions.}
  \label{tab:isr}
\end{table}
With REAPER priors, bigram+ features and 
20 interactions,
the top part of Table \ref{tab:isr} shows a clear advantage to IMP in terms of NDCG@1\%,
which outperforms 
the previous state of the art, BT-UNC 
(significant with $p\ll .01$ 
on all datasets).
In terms of $r$, IMP is out-performed by TP 
(significant with $p\ll .01$ 
on all datasets), 
which appears more balanced between
finding the best candidate and learning the ratings for all candidates.
UNPA improves slightly over random sampling for both metrics,
while EIG is stronger
due to a better focus on epistemic uncertainty.
Unlike IMP, TP does not always outperform EIG
on NDCG@1\%. 

Figure \ref{fig:progress_duc01} shows the progress of each method with 
increasing numbers of interactions on DUC'01. The slow progress of the 
BT baselines is clear, illustrating the advantage the Bayesian methods have
as a basis for active learning by incorporating uncertainty estimates and prior predictions.


\begin{figure}[h]
    \centering
    \includegraphics[width=\textwidth,clip=True,trim=15 60 45 385]{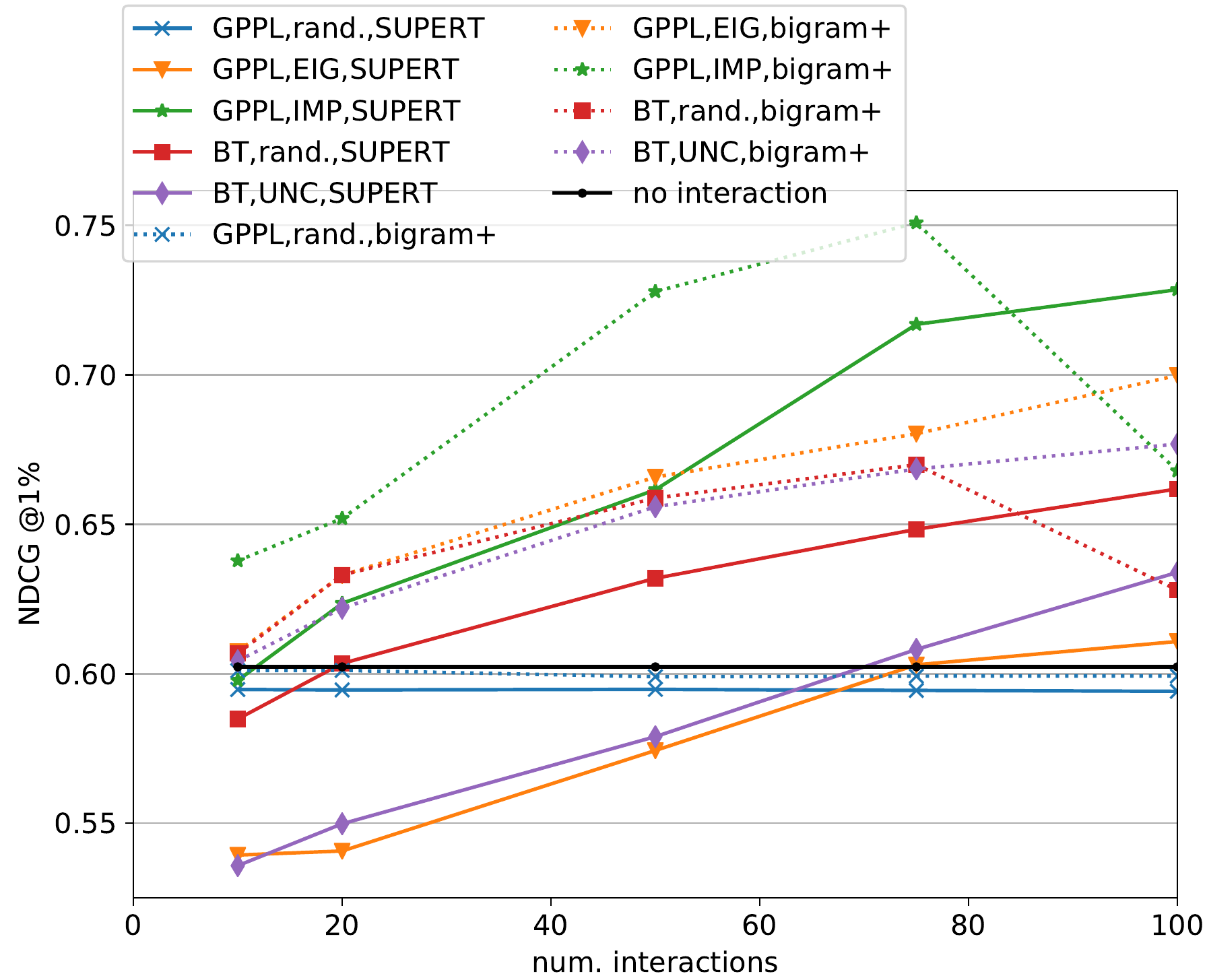}
    \caption{
    DUC'01, SUPERT prior, changes in NDCG@1\% with increasing number of interactions.
        \label{fig:progress_supert}
    }
\end{figure}
The lower part of Table \ref{tab:isr} and Figure \ref{fig:progress_supert} confirm the superior NDCG@1\% scores of
IMP 
with the stronger
SUPERT priors.
However, while pretrained SUPERT outperforms REAPER,
the results after 20 rounds of interaction
with bigram+ features are almost identical,
suggesting that user feedback helps mitigate the weaker
pretrained model.
With only 20 interactions, bigram+ features work better
than SUPERT embeddings as input to our interactive learners,
even with the best-performing method, IMP,
since there are fewer features and the model can cope better with
limited labelled data. With 100 interactions,
SUPERT embeddings provide superior performance as there are sufficient
labels to leverage the richer input embeddings.

\subsection{RL for Summarisation}\label{sec:expts_rl}

\begin{figure}
    \centering
    \includegraphics[width=0.8\columnwidth,clip=false,trim=10 0 20 10]{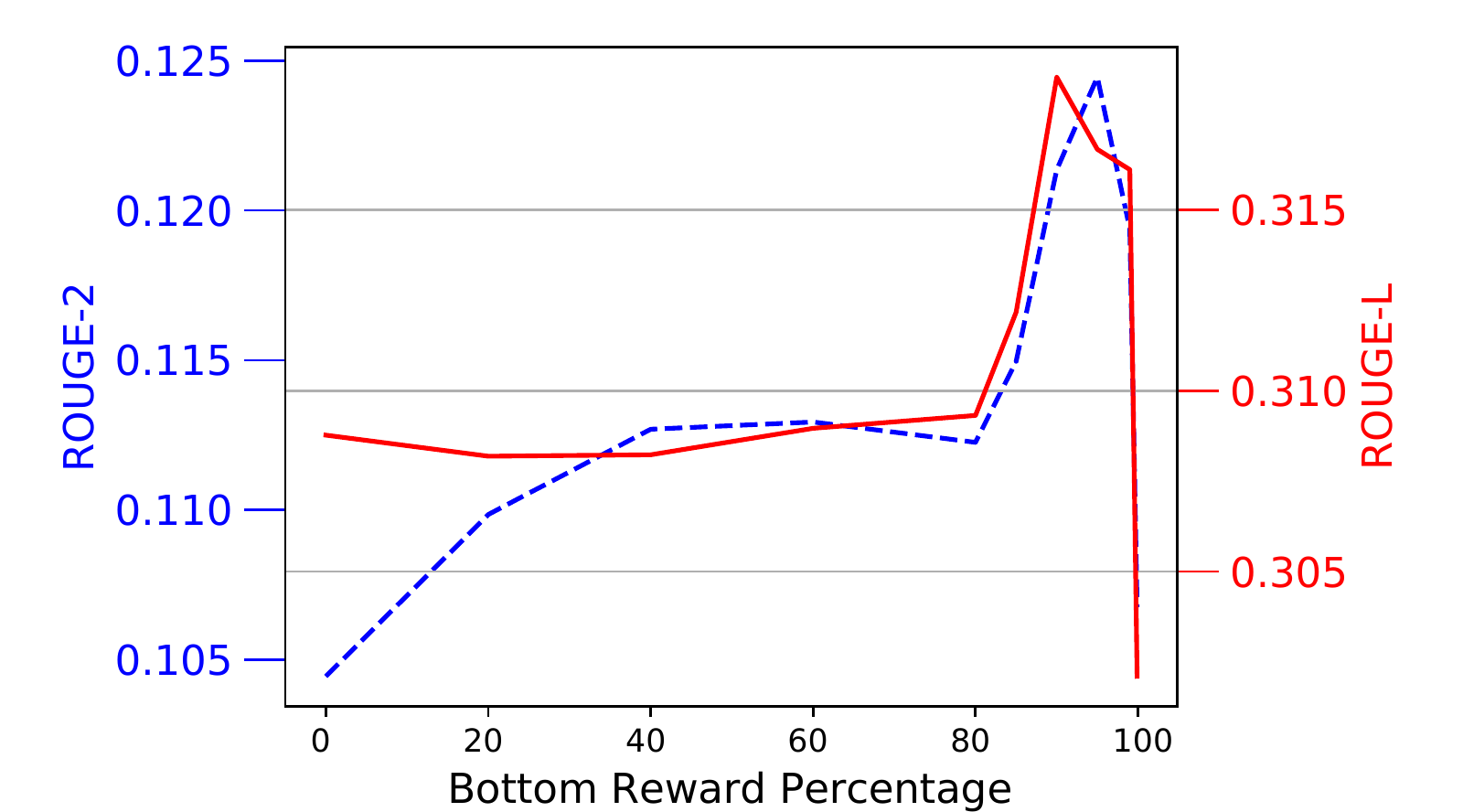}
    \caption{Performance of RL on DUC'01 
    when the rewards for the bottom $x\%$ summaries are flattened to one. Dashed line =
    ROUGE-2, solid line = ROUGE-L.}
    \label{fig:flattened_rewards}
\end{figure}
\begin{table*}[ht]
 \centering
 \small
 \setlength{\tabcolsep}{3pt}
  \begin{tabular}{llllllll@{\hspace{0.3cm}} llll@{\hspace{0.3cm}}llll}
    \toprule
    \#intera & Learner & Features & Stra & \multicolumn{4}{c}{DUC'01} & \multicolumn{4}{c}{DUC'02} & \multicolumn{4}{c}{DUC'04} \\
    -ctions & & & -tegy & R1 & R2 & RL & RSU4 & R1 & R2 & RL & RSU4 & R1 & R2 & RL & RSU4   \\
    \midrule 
    0 & SUPERT & N/A & none &
    .324 & .061 & .252 & .097 & 
    .345 & .070 & .270  & .107 &
    .375 & .086 & .293 & .128 \\
    \midrule
    20 & BT & bigrams+ &  UNC & 
    .335 & .072 & .265 & .104 &
    .364 & .086 & .286 & .120 &
    .390 & .101 & .307 & .136 \\
    20 & GPPL & bigrams+ & rand. &
    .324 & .064 & .252 & .097 &
    .358 & .081 & .281 & .115 &
    .383 & .095 & .302 & .131\\
    20 & GPPL & bigrams+ &EIG & 
    .346 & .073 & .269 & .110 &
    .377 & .095 & .295 & .126 &
    .394 & .106 & .310 & .137 \\
    20 & GPPL & bigrams+ &IMP & 
    \textbf{.355} & \textbf{.086} & \textbf{.277} & \textbf{.114} &
    \textbf{.385} & \textbf{.103} & \textbf{.300} & \textbf{.130} &
    \textbf{.419} & \textbf{.122} & \textbf{.331} & \textbf{.154}\\
    \midrule
    100 & BT & SUPERT & UNC &
    .337 & .072 & .264 & .104 &
    .366 & .086 & .284 & .118 &
    .377 & .090 & .297 & .128 
    \\
    100 & GPPL & SUPERT & rand. & 
   .317 & .057 & .247 & .092 &
   .344 & 071 & .270 & .107 &
   .372 & .087 & .292 & .124 
     \\
    100 & GPPL & SUPERT &EIG &
    .331 & .070 & .259 & .101 &
    .367 & .088 & .287 & .120 &
    .394 & .103 & .309 & .136 \\
    100 & GPPL & SUPERT & IMP & 
    \textbf{.370} & \textbf{.100} & \textbf{.293} & \textbf{.123} &
    \textbf{.406} & \textbf{.118} & \textbf{.316} & \textbf{.140} &
    \textbf{.422} & \textbf{.130} & \textbf{.337} & \textbf{.155}\\ 
    \bottomrule
    \end{tabular}
  \caption{RL for summarisation: ROUGE scores of final summaries, 
  mean over 10 repeats with different random seeds. 
  Once the rewards are fixed, the
  performance of RL is stable:
  standard deviation of each result is $<0.004$.
  }
  \label{tab:rl}
\end{table*}

We now investigate whether our approach also improves performance
when the ranking function is used 
to provide rewards for a reinforcement learner.
Our hypothesis is that it does not matter whether the rewards assigned to
bad candidates are correct, 
as long as they are distinguished from good candidates, 
as this will prevent the policy from choosing bad candidates to present to the user.

To test the hypothesis, we simulate a \emph{flat-bottomed} reward function for summarisation on 
DUC'01:
first, for each topic, we set the rewards for 
the 10,000 sampled summaries (see § \ref{subsec:inter_summary_rating})
to the gold standard, 
$R_{comb}$ (Eq.~\eqref{eq:r_comb}, 
 normalised to $[0,10]$).
Then, we 
set the rewards for a varying percentage
of the lowest-ranked summaries to 1.0 (the flat bottom).
We train the reinforcement learner on the 
flat-bottomed rewards and plot ROUGE scores for the 
proposed summaries in Figure \ref{fig:flattened_rewards}.
The 
performance of the learner 
actually increases as candidate values 
are flattened until around 90\% of the summaries have the same value.
This supports our hypothesis that the user's 
labelling effort should be spent 
on the top candidates.

We now use the ranking functions 
learned in the previous summary rating task as rewards for reinforcement learning.
As examples, we take the rankers
learned using SUPERT priors with
bigram+ features and 20 interactions 
and with SUPERT embeddings and 100 interactions.
We replicate the RL setup of ~\citet{gao2018april}
for interactive multi-document summarisation, which
previously achieved state-of-the-art 
performance using the BT learner with UNC.
The RL agent models the summarisation process as 
follows: there is a current state, represented by
the current draft summary; 
the agent
uses a \emph{policy} to select a sentence to be concatenated
to the current draft summary or to terminate the summary construction.
During the learning process, the agent receives a 
reward after terminating, 
which it uses to 
update its policy 
to maximise these rewards. 
The model is trained for 5,000 episodes
(i.e. generating 5,000 summaries and receiving their 
rewards), then the policy is 
used to produce a summary.
We compare the produced summary 
to a human-generated model summary using ROUGE.
By improving the reward function, we hypothesise 
that the quality of the resulting summary will also improve.

Table \ref{tab:rl} shows that 
the best-performing method from the previous tasks,
IMP, again produces a strong improvement 
over 
the previous state of the art,
BT with UNC (significant with $p\ll 0.01$
in all cases), as well as GPPL with EIG.
With 20 interactions and bigram+ features, EIG
also outperforms BT-UNC, indicating
the benefits of the Bayesian approach, 
but this is less clear with SUPERT embeddings, 
where the high-dimensional embedding space may lead to sparsity problems for the GP. 
The standard deviation in performance over multiple runs of RL is <0.004 for all metrics, datasets, and methods, suggesting that the advantage gained by using IMP is robust to randomness in the RL algorithm.
The results 
confirm that gains in NDCG@1\% made by BO over
uncertainty-based strategies
when learning the utilities translate to better  summaries produced by
reinforcement learning in a downstream task.

\subsection{Limitations of User Simulations}

By testing our interactive process with simulated users, we were able to 
compare numerous methods with a fixed labelling error rate.
The user labels were simulated using data from real individuals: 
the gold answers for cQA were chosen by the user who posed the question,
and the three model summaries for each topic in the DUC datasets
were each authored by a different individual.
While this work shows the promise of BO,
further work is needed to test specific NLP applications
with real end users. 
Our experiments illustrate  
plausible applications where users 
compare texts of up to 100 words
and gain substantial performance advantages.
Other applications require
a broader study of reading and labelling time versus performance benefits and user satisfaction. It may also be possible to select chunks of longer documents for
the user to compare, rather than reading whole documents.

Another dimension to consider is that 
real users may make systematic, rather than random errors.
However, in the applications we foresee,
it is accepted that their preference labels will often
diverge from any established gold standard,
as users adapt models to their own information needs.
 Future work may therefore apply interactive approaches to more subjective
  NLP tasks, such as adapting a summary to 
  more personal information needs.


\section{Conclusions}
\label{sec:conclusion}
We proposed a novel approach to interactive text ranking 
that uses Bayesian optimisation (BO) to identify top-ranked texts by 
acquiring pairwise feedback from a user and applying
Gaussian process preference learning (GPPL).
Our experiments showed  
that BO
significantly improves the accuracy of answers chosen 
in a cQA task with small amounts of feedback, 
and leads to summaries that better match human-generated model summaries
when used to learn a reward function for reinforcement learning.


Of two proposed Bayesian optimisation strategies,
we found that expected improvement (IMP) 
outperforms Thompson sampling (TP) if the goal is to optimise the 
proposed best solution.
TP may require a larger number of interactions due to its
random sampling step.  
IMP is effective in both cQA and summarisation tasks, 
but has the strongest impact on cQA with
only 10 interactions.
This may be due to the greater
sparsity of candidates in cQA (100 versus 10,000 for summarisation), which 
allows them to be more easily discriminated by the model,
given good training examples. 
Further evaluation with real users is required
to gauge the quantity of feedback 
needed in a particular domain.

When using high-dimensional BERT embeddings as inputs, 
GPPL requires more labels to achieve substantial improvements.
Future work may therefore investigate recent dimensionality reduction methods~\citep{raunak-etal-2019-effective}.
We found that performance improves when including prior predictions as the GPPL prior mean but
it is unclear how best to 
estimate confidence in the prior predictions --
here we assume equal confidence in all prior predictions. 
Future work could address this by adapting the GPPL prior covariance matrix to kick-start BO. 
The method is also currently limited to a single set of prior
predictions: in future we intend to 
integrate predictions from several models.



\section*{Acknowledgements}

This work was supported by the German Research Foundation (DFG) 
as part of the EVIDENCE project (grant GU 798/27-1)
and the PEER project (grant GU 798/28-1).
We would like to thank the reviewers and journal editors for their extremely helpful feedback.

\bibliographystyle{acl_natbib}
\bibliography{2019_TACL_GPPRL}

%

\end{document}